# Automating Coreference:
# The Role of Annotated Training Data[1]

### Lynette Hirschman, Patricia Robinson, John Burger, Marc Vilain

The MITRE Corporation
Bedford, MA 01730
{lynette,parann,john,mbv}@mitre.org

## Abstract

We report here on a study of interannotator agreement in the coreference task as defined by the Message Understanding Conference (MUC-6 and MUC-7). Based on feedback from annotators, we clarified and simplified the annotation specification. We then performed an analysis of disagreement among several annotators, concluding that only 16% of the disagreements represented genuine disagreement about coreference; the remainder of the cases were mostly typographical errors or omissions, easily reconciled. Initially, we measured interannotator agreement in the low 80's for precision and recall. To try to improve upon this, we ran several experiments. In our final experiment, we separated the tagging of candidate noun phrases from the linking of actual coreferring expressions. This method shows promise -- interannotator agreement climbed to the low 90s -- but it needs more extensive validation. These results position the research community to broaden the coreference task to multiple languages, and possibly to different kinds of coreference.

## Background

Underlying MITRE's natural language research is the belief – for coreference, as well as other linguistic tasks, such as segmentation, name identification, and syntax – that the key to building adaptive or machine-learned algorithms are corpora of annotated data, coupled with evaluation methods to compare annotations against a "gold standard,". Because of this dependence on (hand) annotated data, systems based on supervised learning can learn only those procedures that humans can perform reliably. If people cannot do a task reliably (e.g., agree on whether one phrase is coreferential with another), then we do not understand the phenomenon well enough to write a program to do it. Thus the preparation of reliable data sets, with good interannotator agreement, is a prerequisite to corpus-based or machine learning research.

Coreference evaluation was introduced as a new domain-independent task at the 6th Message Understanding Conference (MUC-6) in 1995. The MUC-6 (and the upcoming MUC-7) evaluation focuses on a **subset** of coreference, namely coreferring nouns and noun phrases (including proper names) that refer to the same entity (the **IDENTITY** relation). This subset of the coreference task was chosen because there was no consensus at the time on how to broaden the task to include other kinds of coreference (e.g., part/whole relations or extensions to sentential anaphora).

The MUC coreference task is defined in terms of SGML mark-up which consists of a **COREF** tag and a unique identifier for each noun or noun phrase involved in identity coreference, as shown in Fig. 1. For example, **ID=1** is associated with *James J. (Whitey) Bulger*, and **ID=3** with *his*. Phrases that refer back to a previously mentioned entity have an additional **REF** pointer that establishes coreference. For example, the **REF=1** attribute links the phrase *his* with phrase **ID=1**, *James J. (Whitey) Bulger*. The annotation also includes two other attributes, a type attribute (always **TYPE="IDENT"** in the MUC data), and the **MIN** attribute that identifies the minimal (head) element in a coreferring phrase, e.g., **MIN= "winnings"** for the phrase *his lottery winnings*. The **MIN** attribute is used by the scoring algorithm.

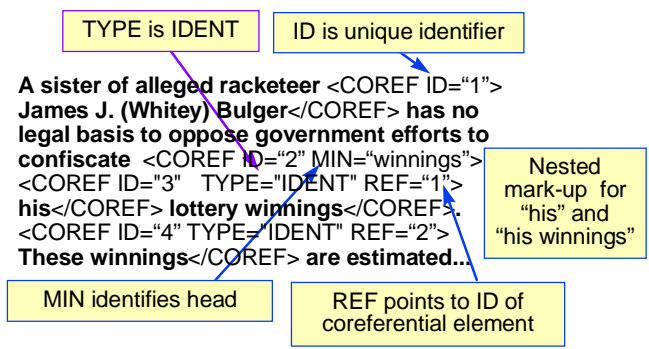

**Figure 1: SGML Mark-up for MUC Coreference**

It is clear from Figure 1 that reading the raw SGML is difficult. To facilitate manual annotation and visualization of coreference annotations, a number of tools have been

---

|   |   | 1 | 2 | 3 | 4 | 5 |
|---|---|---|---|---|---|---|
| 7 | FUGITIVE RACKETEER'S SISTER | **sister** of alleged racketeer James J. (Whitey) Bulger | | | | |
| 11 | | A **sister** of alleged racketeer James J. (Whitey) Bulger | his **sister** Jean Holland | | Holland her | Holland her |
| 1 | FUGITIVE RACKETEER | **James J. (Whitey) Bulger** his | Bulger his Bulger | Bulger | Bulger | brother Bulger his his |
| 3 | LOTTERY WINNINGS | his lottery **winnings** | Bulger's **winnings**, one–sixth of a 1991 $14.3 million jackpot one–sixth of a 1991 $14.3 million jackpot the **winnings** | Bulger's **winnings** they | | |
| 5 | JUDGE | a federal **judge** | | | US District Judge Douglas P. Woodlock he | |
| 18 | | | the Justice Department The Justice Department | | the Justice Department | |
| 35 | | | | | Bulger's **property** | his **property** |

Top: sentence (or paragraph) number
Side: First ID of chain
Minimal text in **emphasis**
* indicates optional coreference

**Figure 2: Tabular Display of Coreference Relations in Document**

developed. For the annotation process, two tools are available: SRA's Discourse Tagging Tool (Aone and Bennett 1995), and MITRE's Alembic Workbench (Day et al 1997).[2] To aid visualization, we have developed an algorithm that converts SGML to HMTL. This allows the coreference relations to be displayed in tabular form, as shown in Fig 2. In this format, each column corresponds to a paragraph (with the document title in the zero-th column). Each row represents a coreference chain – thus all elements in a row refer to the same entity. We see in Fig. 2 that the first two rows both refer to the racketeer's sister. This highlights an error in the human-generated mark-up. The tabular format makes it easy to spot such errors.

## Evaluating Coreference

For machine-learning algorithms and for evaluation of coreference in general, it is necessary to have a scoring algorithm that determines how well two annotated versions of the same text agree in marking coreferring relations. The current coreference scoring algorithm (Vilain et al. 1995) is based on treatment of the identity coreference relation as a symmetric, transitive relationship. Thus if expression A is coreferential with B, then B is coreferential with A. Similarly if A refers to B and B to C, then A and C are also coreferential. Thus coreference relations impose a partition on the space of noun phrases, such that each entity participates in exactly one "coreference chain" or class. The scoring algorithm generates separate scores for recall (the number of correctly identified coreference links over the number of links in the "gold standard" or key) and precision (the number of correctly identified links over the number of links actually generated in the response).

A number of coreference algorithms have been implemented with varying degrees of success. Figure 3 below shows the performance of the seven participating MUC-6 systems on the coreference task (Coreference Task Definition 1995). The high scoring systems achieved a recall in the high 50's to low 60's and a precision in the low 70's. However, the human interannotator agreement (using one annotator as key, and the other as a response) was in the low 80's. This means that systems cannot really improve much before they bump into an uncertainty about the metric itself. It also means that there are serious issues about whether the task is well-defined, given this level of variability among annotators.

---
[2] The Alembic Workbench can be obtained by ftp from the following url: www.mitre.org/resources/centers/advanced_info/g04h/workbench.html

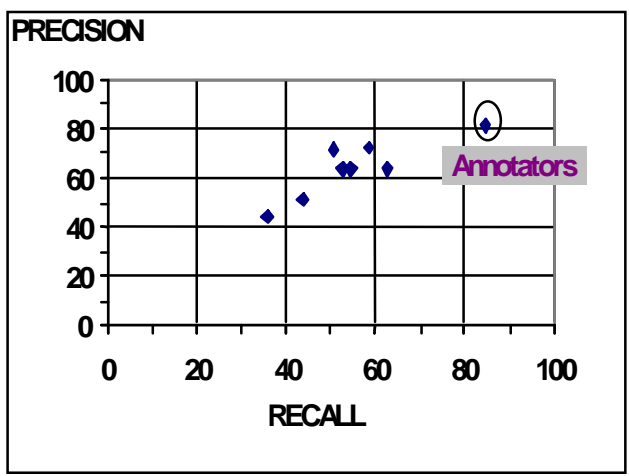

**Figure 3: MUC-6 Coreference Evaluation Results, Including Human Interannotator Agreement**

## Clarifying the Task

When there is interannotator disagreement, it is important to understand the source of the differences. Is the task underdefined or inconsistently defined? If so, the task definition may need to be tightened. Is the task too hard for humans – too many things to remember, too fine a categorization scheme? If so, better tools may improve agreement. In an attempt to understand interannotator differences, we looked both at the consistency of the coreference task definition, and at the specific ways in which annotators disagreed. This led us to refine the coreference task definition.

### Changes to the Task Definition

Feedback from the participants in MUC-6 indicated that change over time was a major source of confusion in the MUC-6 coreference task definition. For example, take the sentences *Mr. Dooner was appointed as CEO in November 1995. He succeeds J. Smith, who served as CEO from 1993 to 1995*. The MUC-6 guidelines were interpreted by some to say that *Dooner* and *CEO* are coreferential in the first sentence; *Smith* and *CEO* are coreferential in the second sentence, and the two instances of *CEO* both refer to the same position in the same company, so they are coreferential. Since the scoring algorithm assumes that identity relations are transitive and symmetric, this means that *Dooner* and *Smith* and both mentions of *CEO* end up in the same coreference equivalence class. This effectively causes the class for *Dooner* and the class for *Smith* to collapse, producing a highly counterintuitive annotation.

To address this, we revised the MUC-7 task definition (MUC-7 Coreference Task Definition 1997) to distinguish <u>extensional</u> mentions of entities (names of individuals) from <u>intensional</u> mentions (reference by description). Intensional mentions, such as *CEO*, are <u>grounded</u> by association with extensional mentions (e.g., *Mr. Dooner*) which prevents the collapse of coreference chains. Thus in our two sentence example, the first instance of *CEO* can no longer be marked coreferential with the second instance which in grounded by coreference to *J. Smith*. The two coreference chains for *Dooner* and *Smith* are thus appropriately kept separate. This change allowed us to capture basic intuitions about what elements should appear in the same coreference equivalence class, without significant modification to the annotation framework or scoring software. This illustrates the need to clarify the task definition to address annotation problems. In addition to this change, there were a number of other minor changes in the MUC-7 Coreference Task Definition to make the annotation task somewhat more internally consistent.

### Analyzing Interannotator Differences

To understand how to improve interannotator agreement, two of the co-authors (Robinson, Hirschman) served as annotators for a new set of five documents. After independently annotating the data, we performed a detailed analysis of where discrepancies occurred. We achieved interannotator agreement of 81% recall, 85% precision.

| Doc Set1 | EASY | | | | TOTAL | MISSING | HARD (Interp) |
|---|---|---|---|---|---|---|---|
| | Pron | Bugs | Zone | Pred | | | |
| 1 | 0 | 6 | 2 | 0 | 8 | 4 | 4 |
| 2 | 6 | 5 | 3 | 0 | 14 | 41 | 7 |
| 3 | 4 | 2 | 0 | 2 | 8 | 25 | 9 |
| 4 | 0 | 2 | 0 | 1 | 3 | 8 | 2 |
| 5 | 1 | 0 | 0 | 5 | 6 | 1 | 0 |
| SUM | 11 | 15 | 5 | 8 | 39 | 79 | 22 |

**Figure 4: Categories of interannotator Differences**

Fig. 4 classifies the differences into three broad categories: "Easy", "Missing" and "Hard" errors, depending on how hard the differences would be to reconcile. The "Hard" errors were places where we genuinely disagreed about the antecedent for a given referring expression. The "Missing" cases were places where one of us failed to label an entire coreference chain; generally, the omissions were dates, or "minor" mentions of things that did not figure prominently in the text and were thus easy to overlook. The "Easy" errors are further subdivided into four categories:
- Pron: failure to mark pronouns;
- Bugs: minor bugs in the coreference scoring algorithm that penalized certain benign differences in

- minimal noun phrase mark-up;
- Zone: failure to mark coreferring expressions that originated in the headline of a text; and
- Pred: failure to mark predicating expressions (e.g., for MUC-6 and MUC-7, in *Dooner is the CEO* we must mark *CEO* as coreferential to *Dooner.*

Overall, 28% of the errors fell into the "Easy" category, 56% into the "Missing" category, and 16% into the "Hard" category. Based on this analysis, we were optimistic that we could cut the interannotator error rate in half by paying attention to these specific kinds of errors.

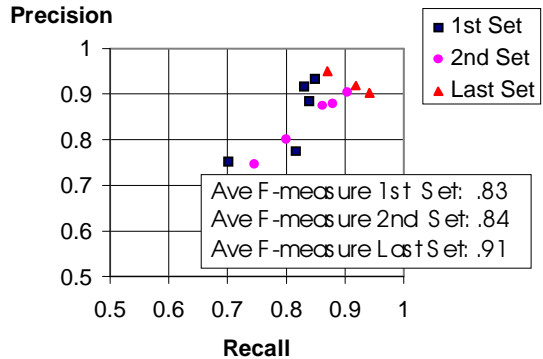

**Figure 5: Interannoator Agreement Experiments**

To test this hypothesis, we chose five new documents and annotated these. We found that our error rate did not decrease significantly. Our interannotator remained the same (84% precision and recall in Fig. 5[3]) after our changes. We decided that we needed to try something different.

Using the approach of the DRAMA specification (Passonneau 1996), we separated annotation into a two-stage process. First, we annotated all "markable" elements, namely any element that was a candidate for participation in a coreference relationship, specifically noun phrases or prenominal modifiers that were coreferential with noun phrases. After identifying all markables in a document (heads only), in a separate pass we then linked the coreferring elements. We found, in three test documents, that this strategy seemed to reduce the error rate (F-measure 91%, see Fig. 5). We believe that this is a promising approach for several reasons:

- We were able to use tools to sort possible coreferring expressions, so that duplicate mentions were easy to find and mark;
- We were able to use tools to sort all markables <u>not</u> participating in a coreference relation; at this stage, for example, unlinked pronouns were easy to find;
- We realized that the subtask of identifying markables was a very different task from the task of identifying coreference; in particular, with improved noun phrase parsing, there is a strong likelihood that identification of markable elements can be largely automated.

## Conclusions

While this exercise used too little data and too few annotators to draw firm conclusions, we believe that it has pointed us in an interesting direction, factoring annotation into two distinct stages: identification of "markable" elements, followed by linkage into coreference chains. This approach should prove useful in building a trainable system. Second, if we can reduce the noise in the training data, as well as the cost of producing the data, we will be able to obtain both more data and cleaner data, which will support the training of more robust and accurate systems. Finally, it is critical that we ensure that we produce annotations that makes sense – when humans do not know how to annotate change over time, for example, then we should not attempt to develop automated systems that reproduce our confusion. We need to rationalize the task definition to ensure that humans can do the task reliably.

These findings position the research community to achieve improved interannotator agreement and to learn better algorithms from the data. Once we understand the issues for one kind of coreference (identity coreference) for one language (English), we can extend the framework to multiple languages, to broader coverage of coreferring expressions, (verb phrase and clause level coreference) and to more types of coreference, such as class-instance or part-whole coreference.

## References

Aone, C.; and Bennett, S. 1995. Evaluating Annotated and Manual Acquisition of Anaphora Resolution Strategies. Proc. of 33rd Annual Meeting of the ACL. 122-129.

Coreference Task Definition, Version 2.3. 1995. *Proc. of the 6th Message Understanding Conf.,* 335-344.

Day, D.; Aberdeen, J.; Hirschman, L.; Kozierok, R.; Robinson, P.; and Vilain, M. 1997. Mixed Initiative Development of Language Processing Systems. *Proc. of the Fifth Conference on Applied Natural Language Processing*. 348-355.

MUC-7 Coreference Task Definition, Version 3.0. 1997. Available from chinchor@gso.saic.com.

Passonneau, R. 1996. Instructions for Applying Discourse Reference Annotation for Multiple Applications (DRAMA). Draft document. Columbia University. Available via ftp at ftp.dfki.de/pub/dri

Vilain, M.; Burger, J.; Aberdeen, J.; Connolly, D.; and Hirschman, L. 1995. A Model Theoretic Coreference Scoring Scheme. *Proc. of the 6th Message Understanding Conf.,* 45-52.

---

[3] Figure 5 gives results in terms of a balanced F-measure, calculated as F = 2*P*R/(P+R).